\documentclass[10pt,twocolumn,letterpaper]{article}

\usepackage{cvpr}
\usepackage{times}
\usepackage{epsfig}
\usepackage{graphicx}
\usepackage[]{subfig}
\usepackage{amsmath}
\usepackage{amssymb}
\usepackage{enumitem}
\usepackage{xcolor}

\usepackage[pagebackref=true,breaklinks=true,letterpaper=true,colorlinks,bookmarks=false]{hyperref}

\cvprfinalcopy 

\ifcvprfinal\fi

\newcommand{\bx}{\mathbf{x}}
\newcommand{\be}{\mathbf{e}}
\newcommand{\bg}{\mathbf{g}}

\newcommand{\bv}{\mathbf{v}}

\begin{document}

\title{Deep Hierarchical Parsing for Semantic Segmentation}

\author{
Abhishek Sharma\\
Computer Science Department\\
University of Maryland\\
{\tt\small bhokaal@cs.umd.edu}
\and
Oncel Tuzel\\
MERL\\
Cambridge\\
{\tt\small oncel@merl.com}
\and
David W. Jacobs\\
Computer Science Department\\
University of Maryland\\
{\tt\small djacobs@umiacs.umd.edu}
}

\maketitle

\begin{abstract}
   This paper proposes a learning-based approach to scene parsing inspired by the deep Recursive Context Propagation Network (RCPN). RCPN is a deep feed-forward neural network that utilizes the contextual information from the entire image, through bottom-up followed by top-down context propagation via random binary parse trees. This improves the feature representation of every super-pixel in the image for better classification into semantic categories. We analyze RCPN and propose two novel contributions to further improve the model. We first analyze the learning of RCPN parameters and discover the presence of bypass error paths in the computation graph of RCPN that can hinder contextual propagation. We propose to tackle this problem by including the classification loss of the internal nodes of the random parse trees in the original RCPN loss function. Secondly, we use an MRF on the parse tree nodes to model the hierarchical dependency present in the output. Both modifications provide performance boosts over the original RCPN and the new system achieves state-of-the-art performance on Stanford Background, SIFT-Flow and Daimler urban datasets.
\end{abstract}


\section{Introduction}
Semantic segmentation refers to the problem of labeling every pixel in an image with the correct semantic category. Handling the immense variability in the appearance of semantic categories requires the use of context to achieve human-level accuracy, as shown, for example, by \cite{torralbaContext,urtasunHuman,alanHuman}.  Specifically, \cite{urtasunHuman,alanHuman} found that human performance in labeling a super-pixel is worse than a computer when both have access to that super-pixel only.  Effectively using context presents a significant challenge, especially when a \emph{real-time} solution is required.

An elegant deep recursive neural network approach for semantic segmentation was proposed in \cite{rcpn}, referred to as RCPN. The main idea was to facilitate the propagation of contextual information from each super-pixel to every other super-pixel through random binary parse trees. First, a {\em semantic mapper} mapped visual features of the super-pixels into a semantic space. This was followed by a recursive combination of semantic features of two adjacent image regions, using a \emph{combiner}, to yield the holistic feature vector of the entire image, termed the root feature. Next, the global information contained in the root feature was disseminated to every super-pixel in the image, using a \emph{decombiner}, followed by classification of each super-pixel via a \emph{categorizer}. The parameters were learned by minimizing the classification loss of the super-pixels by back-propagation through structure \cite{treeProp}. RCPN was shown to outperform recent approaches in terms of per-pixel accuracy (PPA) and mean-class accuracy (MCA). Most interestingly, it was almost two orders of magnitude faster than competing algorithms.

RCPN's speed and state-of-the-art performance motivate us to carefully analyze it.  In this paper we show that it still has some weaknesses and we show how to remedy them.  In particular, the direct path from the semantic mapper to the categorizer gives rise to bypass errors that can cause RCPN to bypass the combiner and decombiner assembly. This can cause back-propogation to reduce RCPN to a simple multi-layer neural network for each super-pixel.  We propose modifications to RCPN that overcome this problem

\begin{enumerate}[noitemsep]
 \item {\bf Pure-node RCPN} - We improve the loss function by adding the classification loss of those internal nodes of the random parse trees that correspond to a single semantic category, referred to as pure-nodes. This serves three purposes. a) It provides more labels for training, which results in better generalization. b) It encourages stronger gradients deep in the network. c) Lastly, it tackles the problem of bypass errors, resulting in better use of contextual information.
\item {\bf Tree MRF RCPN} - Pure-node RCPN also provides us with reliable estimates of the internal node label distributions. We utilize the label distribution of the internal nodes to define a tree-style MRF on the parse tree to model the hierarchical dependency between the nodes.
\end{enumerate}
The resulting architectures provide promising improvements over the previous state-of-the-art on three semantic segmentation datasets: Stanford background \cite{stanData}, SIFT flow \cite{liuSIFT} and Daimler urban \cite{daimlerData}. 

The next section describes some of the related works followed by a brief overview of RCPN in Sec.~\ref{sec:rcpn}. We describe our proposed methods in Sec.~\ref{sec:proposed} followed by experiments in Sec.~\ref{sec:experiment}. Finally, we conclude in Sec.~\ref{sec:conclusion}. 

\section{Related Work}
The previous work on semantic segmentation roughly follows two major themes: learning-based and non-parametric models.

Learning-based models learn the appearance of semantic categories, under various transformations, and the relations among them using parametric models. CRF based image models have been quite successful in jointly modeling the appearance and structure of an image; \cite{stanData,munozSeg,urtasunHuman,alanHuman} use CRFs to combine unary potentials obtained from the visual features of super-pixels with the neighborhood constraints. The differences among these approaches are mainly in terms of the visual features, form of the N-ary potentials and the the CRF modeling. A joint-CRF on multiple levels of an image segmentation hierarchy is formulated in \cite{pylonSeg}.  It achieves better results than a flat-CRF owing to the utilization of higher order contextual information coming in the form of a segmentation hierarchy. Multi-scale convolution neural networks are used in \cite{lecunSegmentation} to learn visual feature extractors from raw-image/label training pairs. It achieved impressive results on various datasets using gPb, purity-cover and CRF on top of the learned features. It was extended in \cite{JMLRseg} by feeding in the per-pixel predicted labels using a CNN classifier to the next stage of the same CNN classifier. However, the propagation structure is not adaptive to the image content and only propagating label information did not improve much over the prior work.

A type of learning based model was proposed in \cite{socherSeg} that aims at learning a mapping from the visual features to a semantic space followed by classification. The semantic mapping is learned by optimizing a structure prediction cost on the ground-truth parse trees of training images with the hope that such a training would embed the visual features in a semantically meaningful space, where classification would be easier. However, our experiments using the code provided by the authors show that semantic space mapping is actually no better than a simple 2-layer neural network on the visual features directly.

Recently, a lot of successful non-parametric approaches for natural scene parsing have been proposed~\cite{lanaSIFT-IJCV,liuSIFT,singhSIFT,eigenSIFT,lanaSIFT2013,adobe}. These approaches are instances of sophisticated template matching to retrieve images that are visually similar to the query, from a database of labeled images. The matching step is followed by super-pixel label transfer from the retrieved images to the query image. Finally, a structured prediction model such as CRF is used to jointly utilize the unary potentials with plausible image models. These approaches differ in terms of the retrieval of candidate images or super-pixels, transfer of label from the retrieved candidates to the query image, and the form of the structured prediction model. These approaches are based on nearest-neighbor retrieval that introduces a critical performance/accuracy trade-off. Theoretically, these approaches can utilize a huge amount of data with ever increasing accuracy. But a very large database would require large retrieval-time, which limits the scalability of these methods.

\section{Background Material}\label{sec:rcpn}
In this section, we provide a brief overview of the RCPN based semantic segmentation framework, please refer to \cite{rcpn} for details. 
\subsection{Overview}
RCPN formulates the problem of semantic segmentation as labeling each super-pixel into desired semantic categories. The complete pipeline starting from the input image to the final pixel-wise labels is shown in Fig.~1. It starts with the super-segmentation of the image followed by the extraction of visual features for each super-pixel; \cite{rcpn} used the Multi-scale CNN \cite{lecunSegmentation} to extract per pixel features that are then averaged over super-pixels. RCPN then constructs random binary parse trees obtained using the adjacency information between super-pixels. The leaf-nodes correspond to the initial super-pixels and successive random merger of two adjacent super-pixels builds the internal nodes up to the root node, which corresponds to the entire image. The super-pixel features along with a parse tree are passed through an assembly of four modules: (\emph{semantic mapper}, \emph{combiner}, \emph{decombiner} and \emph{categorizer}, in order) that  outputs labels for each super-pixel. Multiple random parse trees can be used, both during training and testing. At test time, each parse tree can gives rise to different labels for the same super-pixel, therefore, voting is used to decide the final label. 

{\bf Notation:} Throughout this article - $\bv_i$ denotes visual features of $i^{th}$ super-pixel, $\bx_i$ denotes semantic feature of $i^{th}$ super-pixel and $\tilde{\bx}_i$ denotes enhanced super-pixel features. 

\begin{figure*}\label{fig:flowRCPN}
\centering
      \includegraphics[width=1\textwidth]{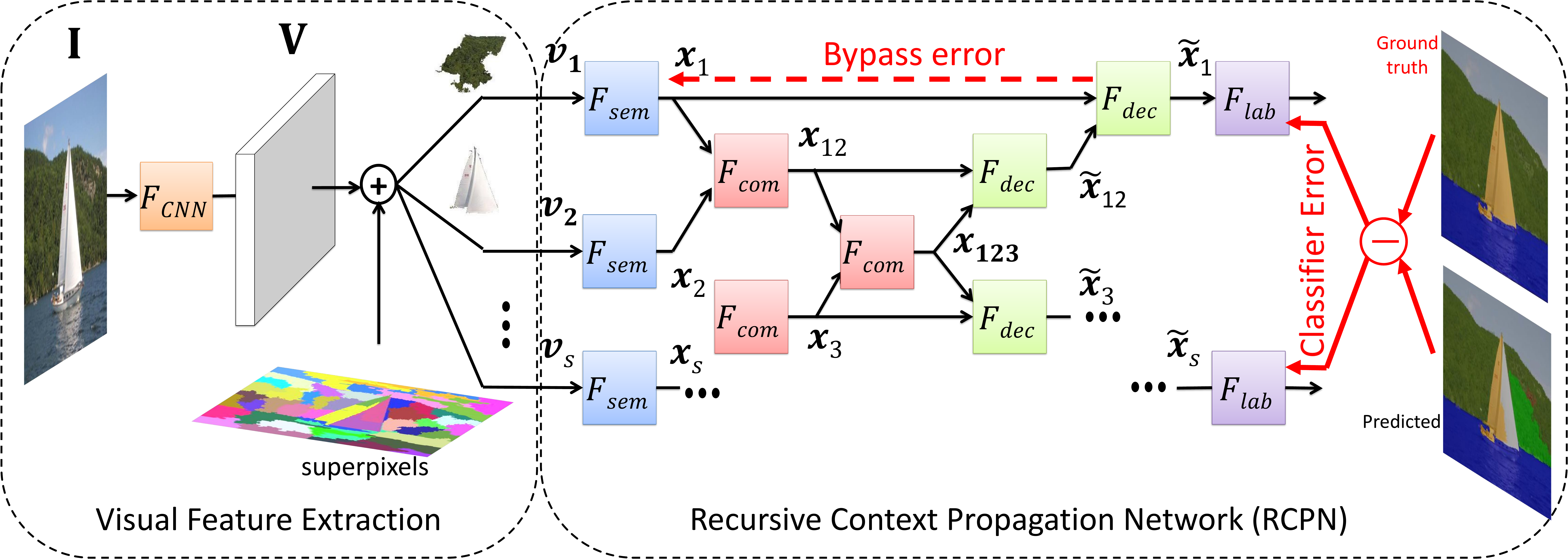}
      \caption{Complete flow diagram of RCPN for semantic segmentation.}
\end{figure*}

{\bf Semantic mapper} is a neural network that maps visual features of each super-pixel to a $d_{sem}$ dimensional semantic feature
\begin{equation} \label{eq:semantic}
\bx_i = F_{sem}(\mathbf{v}_i; W_{sem})
\end{equation}
here, $F_{sem}$ is the network and $W_{sem}$ are the layer weights.

{\bf Combiner: } Combiner is a neural network that recursively maps two child node features ($x_i$ and $x_j$) to their parent feature ($x_{i,j}$). Intuitively, the combiner network attempts to aggregate the semantic content of the children features such that the parent's features become representative of the children. The root features represent the entire image.
\begin{equation} \label{eq:combiner}
	\bx_{i,j} = F_{com}([\bx_{i}, \bx_{j}] ; W_{com}).
\end{equation}
here, $F_{com}$ is the network and $W_{com}$ are the layer weights. 

{\bf Decombiner} is a neural network that recursively disseminates the context information from a parent node to its children through the parse tree. This network maps the semantic features of the child node and its parent to the contextually enhanced feature of the child node. This top-down contextual propagation starts from the root feature and the decombiner is applied recursively up to the enhanced super-pixel features. Therefore, it is expected that every super-pixel feature contains the contextual information aggregated from the entire image.
\begin{equation} \label{eq:decombiner}
	\tilde{\bx}_{i} = F_{dec}([ \bx_{i}, \tilde{\bx}_{i,j}] ; W_{dec}).
\end{equation}
here, $F_{dec}$ is the network and $W_{dec}$ are the layer weights.

{\bf Categorizer } is the final network, which maps the context enhanced semantic features ($\tilde{\bx}_i$) of each super-pixel to one of the semantic category labels; it is a Softmax classifier
\begin{equation}
\mathbf{y}_j = F_{cat}(\tilde{\bx}_i; W_{cat}).
\end{equation}

Together, all the parameters of RCPN are denoted as $W_{rcpn} = \{W_{sem},W_{com},W_{dec},W_{cat}\}$. Let's assume there are $S$ super-pixels in an image $I$ and denote a set of $R$ random parse trees of $I$ as $\mathcal{T}$. Then, the loss function for $I$ is
\begin{equation}\label{eq:loss}
\mathcal{L}(I) = \frac{1}{RS} \sum_{r=1}^{R}{ \sum_{s=1}^{S_i} { L(\mathbf{y}_{r,s},t_s;\mathcal{T}_r,W_{rcpn})  }}
\end{equation}
here, $\mathbf{y}_{r,s}$ is the predicted class-probability vector and $t_s$ is the ground-truth label for the $s^{th}$ super-pixel for random parse tree $\mathcal{T}_r$ and $L(\mathbf{y}_s,t)$ is the cross-entropy loss function. Network parameters, $W_{rcpn}$, are learned by minimizing $\mathcal{L}(I)$ for all the images in the training data.

\section{Proposed Approach}\label{sec:proposed}
In this section, we study the RCPN model, discover potential problems with parameter learning and propose useful modifications to the learning and the model. Our first modifications tackle a potential pitfall during training that stems from the special architecture of RCPN and can reduce it to a simple multi-layer NN. The second modification extends the model by building an MRF on top of the parse trees to utilize the hierarchical dependency between the nodes.

\subsection{Pure-node RCPN}
Here we propose a model that will handle bypass errors.  At the same time, this model solves a problem of gradient attenuation, and also multiplies the training data. For the ease of understanding all our discussions will be limited to 1-layer modules. This result in each of the $W_{sem}$, $W_{com}$, $W_{dec}$ and $W_{cat}$ as matrices.  
Like most deep networks, RCPN also suffers from vanishing gradients for the lower layers. This stems from the vanishing error signal, because the gradient ($\mathbf{g}_l$) for the $l^{th}$ layer depends on the error signal ($\mathbf{e}_{l+1}$) from the layer above -
\begin{equation}
 \mathbf{g}_l = \mathbf{e}_{l+1}  \bx_l^T
\end{equation}
here, $\bx_l$ is the input to the $l^{th}$ layer. For RCPN, vanishing gradients are more of a problem because of very deep parse trees due to recursion. For instance, a 100 super-pixel image will lead to a minimum of ($ log_2(100) \times 2 + 2 > 14$) layers under the strong assumption of perfectly balanced binary parse trees. In practice, we can only create roughly balanced binary trees that often lead to $\sim 30$ layers. 

We show that the internal nodes of the parse tree can be used to alleviate these problem. Each node in the parse tree corresponds to a connected region in the image. The leaf nodes correspond to the initial super-pixels and the internal nodes correspond to the merger of two or more connected regions, referred to as merged-region. We use the term \emph{pure nodes} to refer to the internal nodes of the parse tree associated with the merger of two or more regions of the same semantic category. Therefore, the merged-regions corresponding to the pure nodes can serve as additional labeled samples during training. We empirically found that roughly $65\%$ of all the internal nodes are pure-nodes for all three datasets. We include the classification loss of the pure-nodes in the loss function (Eqn.~\ref{eq:loss}) for training and refer to the new procedure as \emph{pure-node RCPN} or PN-RCPN for short. The classification loss, $\mathcal{L}^{p}(I)$, now becomes -
\begin{equation}\label{eq:pn-loss}
\mathcal{L}^{p}(I) = \mathcal{L}(I) +  \frac{1}{\sum{P_r}} \sum_{r=1}^{R}{ \sum_{p=1}^{P_r} L(\mathbf{y}_{r,p},t_{r,p};\mathcal{T}_r,W_{rcpn})}
\end{equation}
here, $P_r$ is the number of pure-nodes for the $r^{th}$ random parse tree $\mathcal{T}_r$ and subscripts $(r,p)$ map to the $p^{th}$ pure-node for the $r^{th}$ random parse tree. Note that different parse trees for the same image can have different pure nodes.

In order to understand the benefits of PN-RCPN and contrast it with RCPN, we make use of an illustrative example depicted with the help of Fig. 2. The left-half of a random parse tree for an image $I$ with 5 super-pixels, annotated with various variables involved during one forward-backward propagation through RCPN are PN-RCPN are shown in Fig. \ref{fig:bypass} and \ref{fig:bypassPNRCPN}, respectively. We denote, $\mathbf{e}^{cat}_i$ (a $C \times 1$ vector) as the error at enhanced super-pixel nodes; $\mathbf{e}^{dec}_{k}$ (a $2d_{sem} \times 1$ vector) as the error at the decombiner; $\mathbf{e}^{com}_{k}$ (a $2d_{sem} \times 1$ vector) as the error at the combiner and $\mathbf{e}^{sem}_i$ (a $d_{sem} \times 1$ vector) as the error at the semantic mapper. Subscripts $bp$ and $total$ indicate bypass and the sum total error at a node, respectively. We assume a non-zero categorizer error signal for the first super-pixel only, ie $\mathbf{e}^{cat}_{i\neq1} = \mathbf{0}$.  These assumptions facilitate easier back-propagation tracking through the parse tree, but the conclusions drawn will hold for general cases as well. 

The first obvious benefit of using pure-nodes is more labeled samples from the same training data that can improve generalization. The second advantage of PN-RCPN can be understood by contrasting the  back-propagation signals for a sample image for RCPN and PN-RCPN, with the help of Fig. \ref{fig:bypass} (RCPN) and \ref{fig:bypassPNRCPN} (PN-RCPN). Note that in the case of RCPN, the back-propagated training signal was generated at the enhanced leaf-node features and progressively attenuates as it back-propagates through the parse tree, shown with the help of variable thickness solid red arrows. On the other hand, pure-node RCPN has an internal node (shown as a green color node) that injects a strong error signal deep into the parse tree, resulting in stronger gradients even in the deeper layers. Moreover, PN-RCPN \emph{explicitly} forces the combiner to learn meaningful combination of two super-pixels, because incorrect classification of the combined features is penalized.

Now, we come to the third benefit of the PN-RCPN architecture. In what follows, we describe a subtle yet potentially serious problem related to RCPN learning, provide empirical evidence that this problem exists, and argue that PN-RCPN can offer a solution to this problem.

\subsubsection{Understanding the Bypass Error}
During the minimization of the loss functions (Eqn.~\ref{eq:loss} or \ref{eq:pn-loss}), typically, more effective parameters in bringing down the objective function receive stronger gradients and reach their stable state early. Due to the presence of multiple layers of non-linearities and complex connections, the loss function is highly non-convex and the solution inevitably converges to a local minimum. It was shown in \cite{rcpn} that the combiner and decombiner assembly is the most important constituent of the RCPN model. Therefore, we expect the learning process to pay more attention to $W_{com}$ and $W_{dec}$. Unfortunately, the RCPN architecture introduces short-cut paths in the computation graph from the semantic mapper to the categorizer during the forward propagation that gives rise to \emph{bypass errors} during back-propagation. Bypass errors severely affect the learning by reducing the effect of the combiner on the overall loss function, thereby favoring a non-desirable local minimum.

In order to understand the effect of bypass error, we again make use of the example in Fig. 2 to show that bypass paths allow the back-propagated error signals from the categorizer ($\mathbf{e}^{cat}_i$) to reach the semantic mapper through one layer only. On the other hand, $\mathbf{e}^{cat}_i$ goes through multiple layers before reaching the combiner. Therefore, the gradient $g_{com}$ for the combiner is weaker than the gradient for the semantic mapper ($g_{sem}$). 


\begin{figure}
\centering
  \mbox{
    \subfloat[]{
      \label{fig:bypass}
      \includegraphics[width=3.75cm]{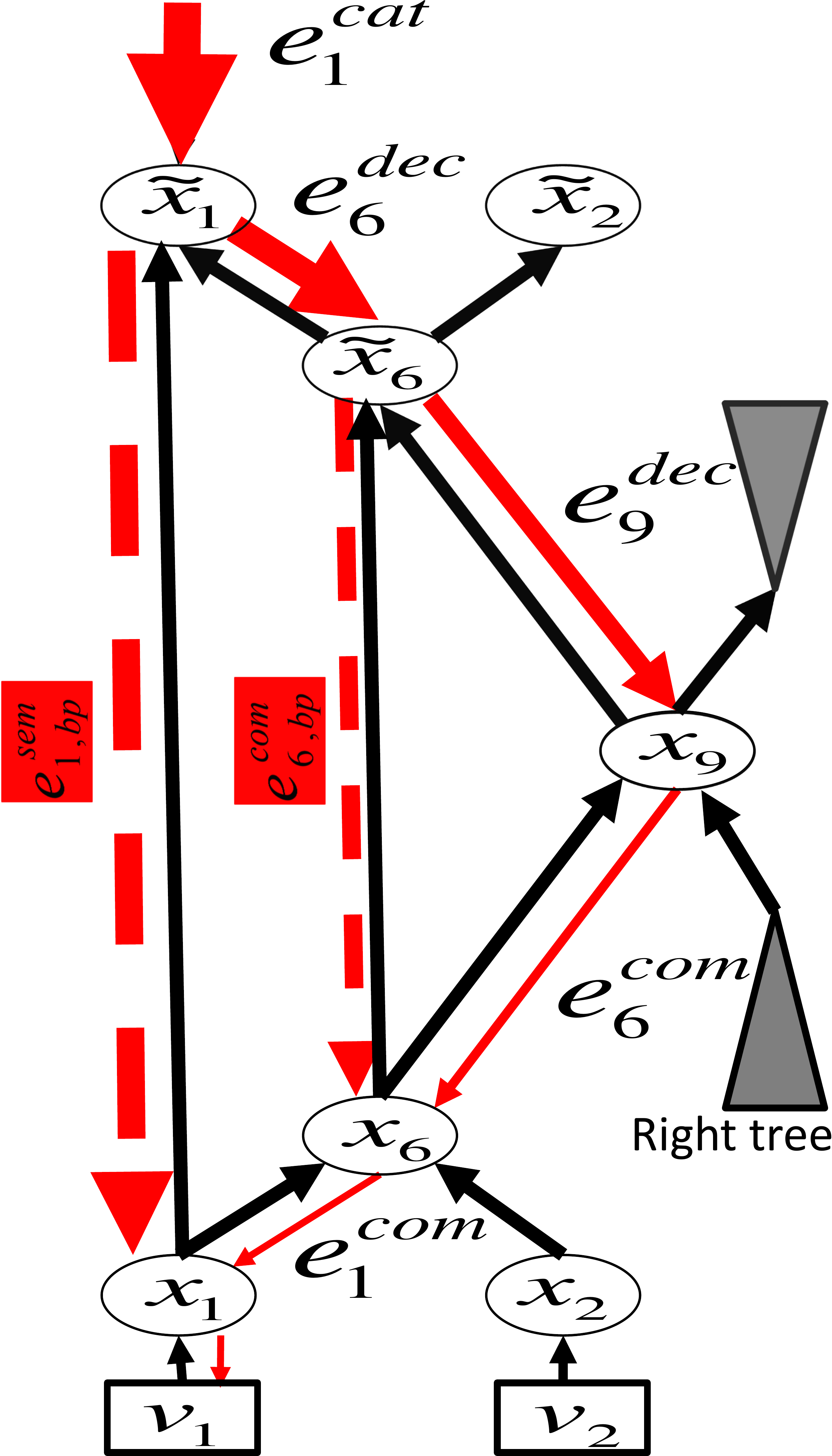}

    }
  } \hfill
  \mbox{
    \subfloat[]{
      \label{fig:bypassPNRCPN}
      \includegraphics[width=3.75cm]{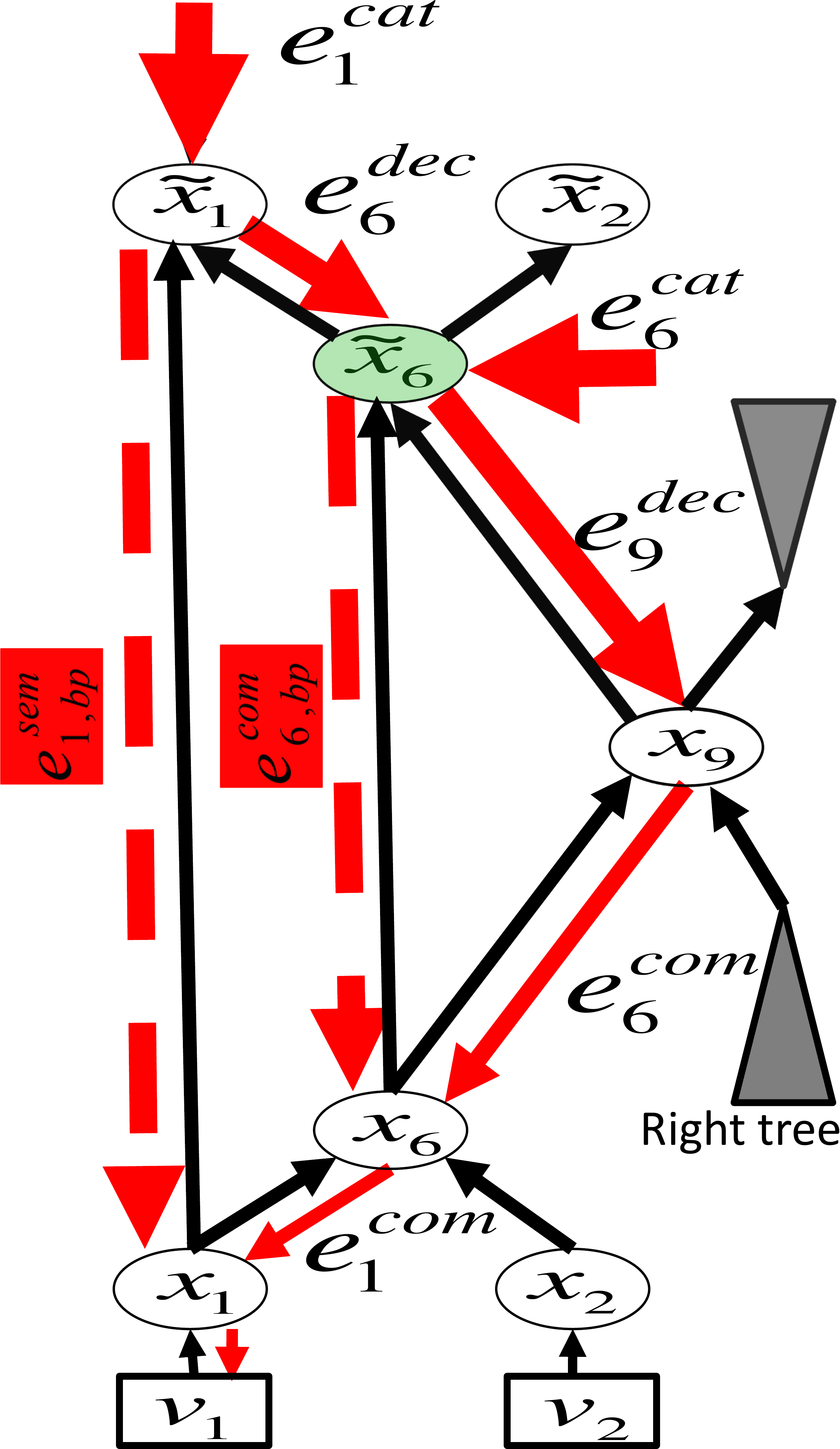}
    }
  } 
  \label{fig:errorProp}
  \caption{Back-propagated error tracking to visualize the effect of bypass error. The variables follow the notation introduces in Sec.~\ref{sec:rcpn}. Forward propagation and back-propagation are shown by solid black and red arrows, respectively. The attenuation of the error signal is shown by variable {\bf width} red arrows. The bypass errors are shown with dashed red arrows. (a) RCPN: Error signal from $\tilde{\bx}_1$ reaches to $\bx_1$ in just one step, through the bypass path. (b) PN-RCPN introduces pure-nodes classification loss (for $\mathbf{\tilde{x}}_6$), thereby, forcing the network to learn meaningful internal node representation via combiner, thereby, promoting effective contextual propagation.}
\end{figure}
From the Fig.~\ref{fig:bypass} we can see that there are two possible paths for $\mathbf{e}_1^{cat}$ to reach the combiner. One of them requires 2 layers ($\tilde{\bx_1} \rightarrow \tilde{\bx_6} \rightarrow \bx_6$) and the other requires 3 layers ($\tilde{\bx_1} \rightarrow \tilde{\bx_6} \rightarrow \bx_9 \rightarrow \bx_6$). Similarly, $\be_1^{cat}$ can reach $\bx_1$ through a 1 layer bypass path ($\tilde{\bx}_1 \rightarrow \bx_1 $) or a several layers path through the parse tree. Due to gradient attenuation, the smaller the number of layers the stronger the back-propagated signal, therefore, bypass errors lead to $g_{sem} \ge g_{com}$. This can potentially render the combiner network inoperative and guide the training towards a network that effectively consists of a $N_{sem} + N_{dec} + N_{cat}$ layer network from the visual feature ( $\mathbf{v}_i$) to the super-pixel label ($y_i$). This results in little or no contextual information exchange between the super-pixels. In the worst case $W_{dec} = [W ~~ 0]$; this removes the effect of parents on their children features during top-down contextual propagation through the decombiner, thereby completely removing the affect of the combiner from RCPN. Practically, the random initialization of the parameters ensures that they will not converge to such a pathological solution. However, we show that a better local minimum can be achieved by tackling the bypass errors.

\begin{figure}
\centering
  \mbox{
    \subfloat[]{
      \label{fig:gradRCPN}
      \includegraphics[width=6cm]{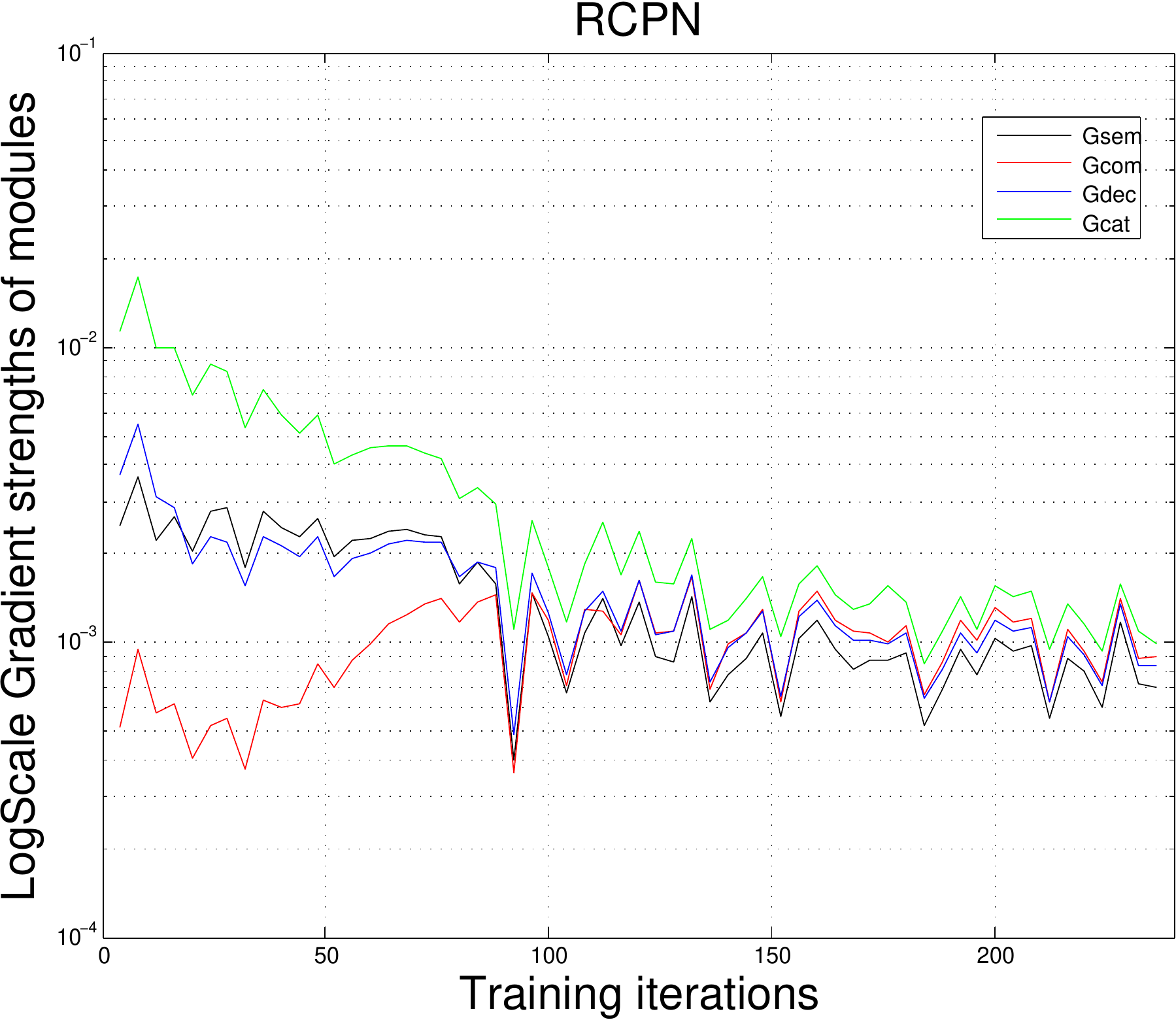}
    }
  }
  \mbox{
    \subfloat[]{
      \label{fig:gradPNRCPN}
      \includegraphics[width=6cm]{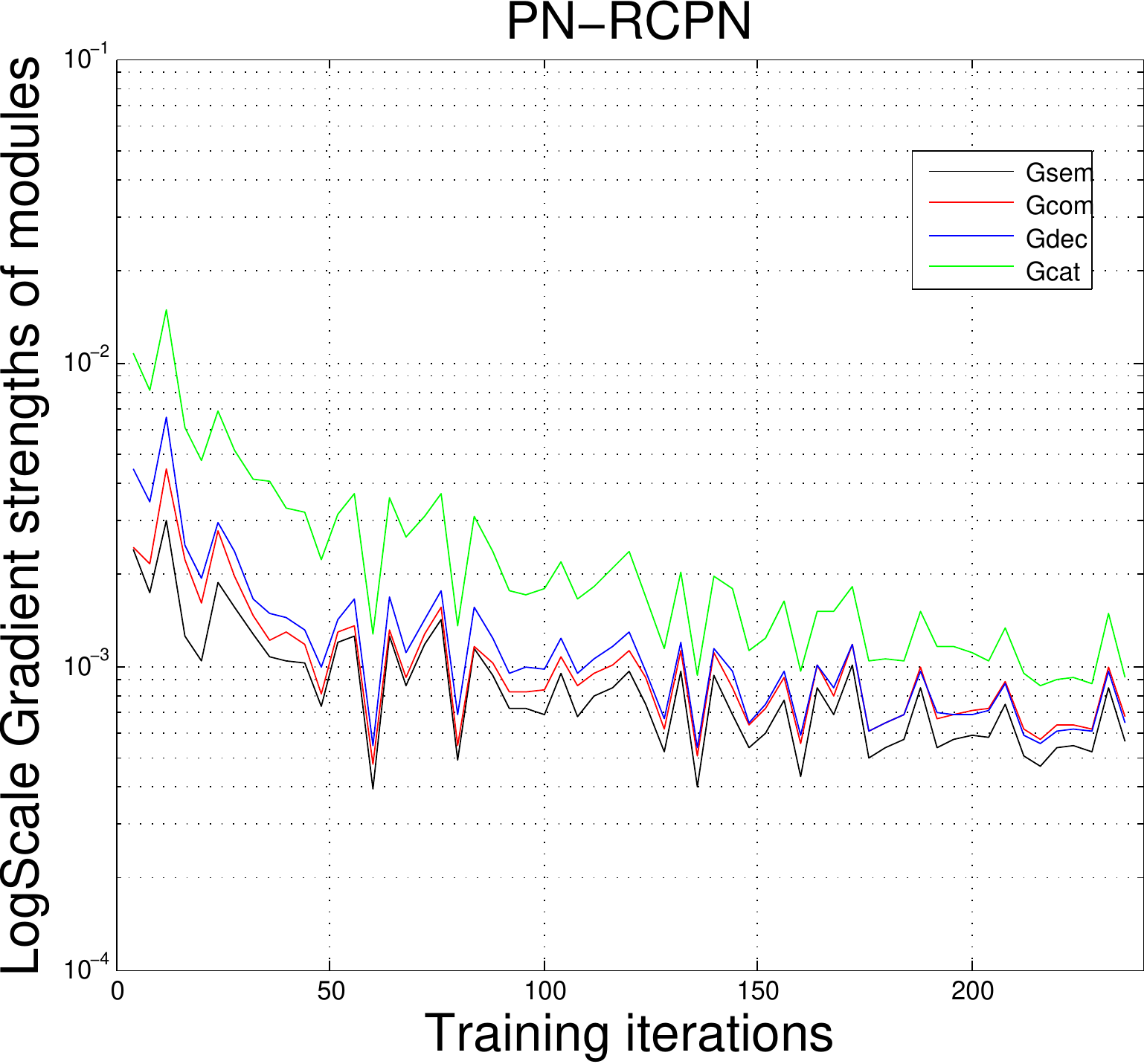}
    }
  }
  \caption{Comparison of gradient strengths of different modules of (a) RCPN and (b) PN-RCPN during training.}
\end{figure}

In order to see that $\bg_{sem} \ge \bg_{com}$, we compute the gradient strengths of each module ($g_{sem}$, $g_{com}$, $g_{dec}$, $g_{cat}$) during training. The gradient strengths of different modules for RCPN and PN-RCPN are normalized by the number of parameters and plotted in Fig.~\ref{fig:gradRCPN} and Fig.~\ref{fig:gradPNRCPN}, respectively. As expected, $g_{cat}$ is the strongest, because it is closest to the initial error signal. Surprisingly, for RCPN $g_{sem}$ is slightly stronger than $g_{dec}$ and significantly stronger than $g_{com}$ during the initial phase of training. Normally, we would expect $g_{sem}$, which is the farthest away from the error signal, to be the weakest due to vanishing gradients. This observation suggests that the initial training phase favors a multi-layer NN. However, we also observe that during the later stages of training, $g_{com}$ is comparable to other gradients. Unfortunately, it has been conclusively established, by many empirical studies, that the initial phase of training is crucial for determining the final values of the network parameters, and thereby their performance \cite{pretrain}. From the figure we see that the combiner catches up with the other modules during later stages of training, but by then the parameters are already in the attraction basin of a poor solution. 

On the other hand, the gradients for PN-RCPN (Fig~\ref{fig:gradPNRCPN}) follow the natural order of strength, which gives more importance to the combiner and decombiner than the semantic mapper during the initial training. Fig.~\ref{fig:bypassPNRCPN} provides an intuitive explanation by showing the categorizer error signal ($\be_6^{cat}$) for $\tilde{\bx}_6$ that reaches to the combiner through one layer only ($\textcolor{red}{\mathbf{e}^{com}_{6,bp}}$). To further investigate which of the three aforementioned benefits play the biggest role in improving the performance of PN-RPCN over RPCN, we trained PN-RCPN on SIFT flow under the same setting as Table \ref{tab:SIFT-table}, but we removed as many leaf node labels from the classification loss as the number of pure-nodes. This makes the number of labeled samples equal in both RCPN and PN-RCPN, but leaf-nodes are replaced with pure-nodes. As expected, it still improves PPA and MCA score for PN-RCPN (80.5\% and 35.3\%) vs. RCPN (79.6\% and 33.6\%). This last experiment confirms that inclusion of pure-nodes does not only provide more samples but also helps in overcoming the discussed shortcomings of RCPN.

\subsection{Tree MRF Inference}

\begin{figure}[tb]
\begin{center}
    \includegraphics[trim = 1.0in 3.2in 5.0in 1.0in, clip,height=0.6\linewidth,width=0.6\linewidth]{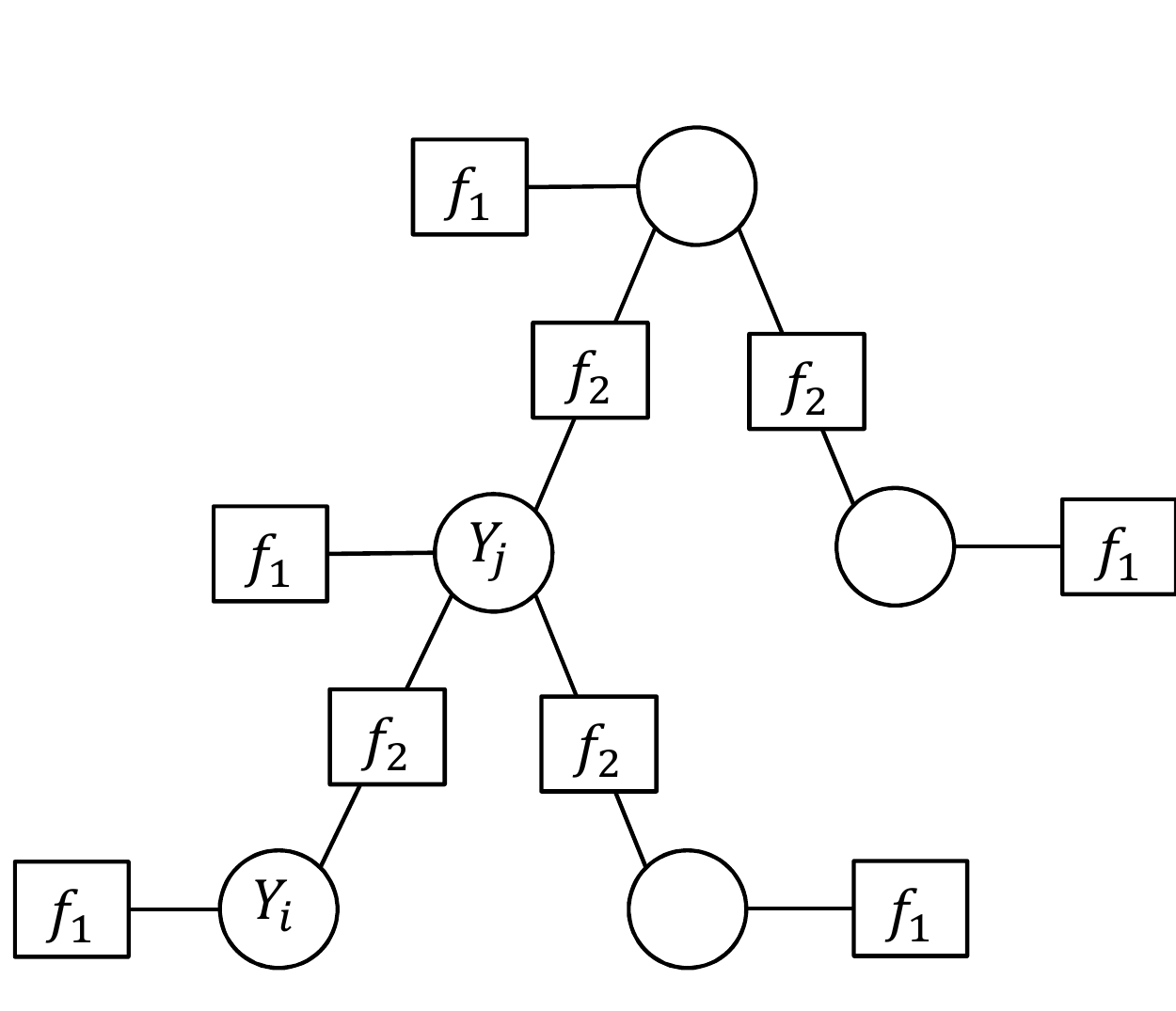}
\end{center}
\caption{Factor graph representation of the MRF model.}
\label{Tree:Factor}
\end{figure}

The pure node extension of RCPN provides the label distributions over merged-regions associated with the internal nodes in addition to individual super-pixel labels. In this section, we describe a Markov Random Field (MRF) structure to model the output label dependencies of the super-pixels while leveraging the internal node label distributions for hierarchical consistency. The proposed MRF uses the same trees structure as that of the parse trees used for RCPN inference.  
A factor graph representation of this MRF is shown in Figure~\ref{Tree:Factor}. 
The variables $Y_i$ are $L$-dimensional binary label vectors associated with each region of the image, $L$ is the number of possible labels. The $k^{th}$ dimension of $Y_i$ is set according to the presence (1) or absence (0) of the $k^{th}$ class super-pixel in the region. 

The unary potentials $f_1$ are given by the label distributions predicted by the RCPN and defined as - 
\begin{equation}
f_1(Y_i) = \frac{-\mathbf{Y_i}^T \log(\mathbf{p}_i)}{\|Y_i\|_1}  
\end{equation}
where $\mathbf{p}_i$ is the softmax output of the categorizer network for super-pixel $i$. If the probabilities given by RCPN are not degenerate, the unary potential prefers to assign a single label, that of the node with the highest probability.

The pairwise potentials $f_2$ are introduced to impose consistency between a pair of child and parent regions. The parent region \emph{must} include all the labels assigned to its children regions, which is a hard constraint:
\begin{equation}
f_2(Y_i, Y_j) = \begin{cases}
    \infty, & \text{if $\mathcal{S}(Y_i) \setminus \mathcal{S}(Y_j) \neq \emptyset$}.\\
    0, & \text{otherwise}.
  \end{cases}
\end{equation}
where node $j$ is the parent node of $i$ and $\mathcal{S}(Y)$ is the set of all the labels in the merged-region with label vector $Y$.

The unary potentials $f_1$ utilize all levels of the tree simultaneously and prefer purer nodes, whereas pairwise potentials, $f_2$ enforce consistency across the tree hierarchy. This design allows for spatial smoothness at lower levels and mixed labeling at the higher levels. The tree structure of the MRF affords exact decoding using max-product belief propagation. The size of the state space is exponential in the number of labels. However, in practice there are rarely more than a handfull of different object classes within an image. Therefore, to reduce the size of the state space, we first identify different labels predicted by the RCPN and only retain the $9$ most frequently occurring super-pixel labels per image. 

\section{Experimental analysis}\label{sec:experiment}
In this section we evaluate the performance of proposed methods for semantic segmentation on three different datasets: Stanford Background, SIFT Flow and Daimler Urban. Stanford background dataset contains 715 color images of outdoor scenes, it has 8 classes and the images are approximately $240 \times 320$ pixels. We used the 572 train and 143 test image split provided by \cite{socherSeg} for reporting the results. SIFT Flow contains 2688, $256 \times 256$ color images with 33 semantic classes. We experimented with the train/test (2488/200) split provided by the authors of \cite{lanaSIFT-IJCV}. Daimler Urban dataset has 500, $400 \times 1024$ images captured from a moving car in a city, it has 5 semantic classes. We trained the model using 300 images and tested on the rest of the 200 images, the same split-ratio has been used by previous work on this dataset.

\subsection{Visual feature extraction}
We use a Multi-scale convolution neural network (Multi-scale CNN) \cite{lecunSegmentation} to extract pixel-wise features using publicly available library Caffe \cite{caffe}. We follow \cite{rcpn} and use the same CNN structure with similar preprocessing (subtracting 0.5 from each channel at each pixel location in the RGB color space) at 3 different scales (1,1/2 and 1/4) to obtain the visual features. The CNN architecture has three convolutional stages with $8 \times 8 \times 16 ~conv \rightarrow 2 \times 2 ~ maxpool \rightarrow 7 \times 7 \times 64 ~ conv \rightarrow 2 \times 2 ~ maxpool \rightarrow 7 \times 7 \times 256 ~ conv$ configuration, each max-pooling is non-overlapping. Therefore, every image scale gives a 256 dimensional output map. The outputs from each scale are concatenated to get the final feature map. Note that the $256 \times 3 = 768$ dimensional concatenated output feature map is still 1/4th of the height and width of the input image due to the max-pooling operations. In order to obtain the input size per-pixel feature map we simply scale-up each feature map by a factor of 4 in height and width using Bilinear interpolation. 

We use the publicly available implementation of \cite{mingSeg} to obtain 100 (same as RCPN) and 800 super-pixels per image for SIFT Flow and Daimler Urban, respectively. Daimler uses more super-pixels due to its larger size. For Stanford background, we have used the super-pixels provided by \cite{socherSeg}. 

\subsection{Model Selection}
Unlike most of the previous works that rely on careful hand-tuning and expert knowledge for setting the model parameters, we only need to set one parameter, namely $d_{sem}$, after we have fixed the modules to be 1-layer neural networks. This affords a generic approach to semantic segmentation that can be easily trained on different datasets. For the sake of strict comparison with the original RCPN architecture, we also use 1-layer modules with $d_{sem} = 60$ in all our experiments. \emph{Plain-NN} refers to training a 2-layer NN with 60 hidden nodes, on top of visual features for each super-pixel. \emph{RCPN} refers to the original RCPN model \cite{rcpn}. \emph{PN-RCPN} refers to pure-node RCPN and \emph{TM-RCPN} refers to tree-MRF RCPN.

\subsection{Evaluation metrics}
We have used four standard evaluation metrics  -
\begin{itemize}[noitemsep]
 \item \textbf{Per pixel accuracy (PPA): } Ratio of the correct pixels to the total pixels in the test images, while ignoring the background.
 \item \textbf{Mean class accuracy (MCA): } Mean of the category wise pixel accuracy.
 \item \textbf{Intersection over Union (IoU): } Ratio of true positives to the sum of true positive, false positive and false negative, averaged over all classes. This is a popular measure for semantic segmentation of objects because it penalizes both over- and under-segmentation.
 \item \textbf{Time per image (TPI): } Time required to label an image on GPU and CPU.
\end{itemize}

The results from previous works are taken directly from the published articles. Some of the previous works do not report all four evaluation metrics; we leave the corresponding entry blank in the comparison tables.

\subsection{Stanford Background}
We report our results with CNN features extracted from the original scale only, because multi-scale CNN features overfit, perhaps due to small training data, as observed in \cite{rcpn}. We use 10 and 40 random trees for training and testing, respectively. The results are shown in Table~\ref{tab:stan-table}. From the comparison, it is clear that our proposed approaches outperform previous methods. We observe that PN-RCPN significantly improves the results in terms of MCA and IoU over RCPN. We observe a marginal improvement offered by TM-RCPN over PN-RCPN.

\begin{table}[tb]
  \caption{Stanford background result.}
  \label{tab:stan-table}
  \centering
  \begin{tabular}{|c|c|c|c|c|}
    \hline
    \textbf{Method} & {\bf PPA} & {\bf MCA} & {\bf IoU} & \begin{tabular}{@{}c@{}}{\bf TPI (s)} \\ CPU/GPU\end{tabular}\\
    \hline\hline
    Gould, \cite{stanData}& 76.4 & NA & NA & 30 -- 600 / NA\\
    Munoz, \cite{munozSeg}& 76.9 & NA & NA & 12 / NA\\
    Tighe, \cite{lanaSIFT-IJCV}& 77.5 & NA & NA & 4 / NA  \\
    Kumar, \cite{kumarSeg}& 79.4 & NA & NA &$\leq 600$ / NA  \\
    Socher, \cite{socherSeg}& 78.1 & NA & NA & NA / NA  \\
    Lempitzky, \cite{pylonSeg}& 81.9 & 72.4 & NA & $\geq 60$ / NA  \\
    Singh, \cite{singhSIFT}& 74.1 & 62.2 & NA & 20 / NA \\
    Farabet, \cite{lecunSegmentation} & 81.4 & 76.0 & NA & 60.5 / NA\\
    Eigen, \cite{eigenSIFT} & 75.3 & 66.5 & NA & 16.6 / NA  \\
    Pinheiro, \cite{JMLRseg} & 80.2 & 69.9 & NA & 10 / NA  \\
    \hline \hline
    Plain-NN & 80.1 & 69.7 & 56.4 & 1.1/0.4  \\
    \hline \hline
    RCPN \cite{rcpn} & 81.8 & 73.9 & 61.3 & 1.1/0.4 \\
    \hline \hline
    PN-RCPN & 82.1 & 79.0 & 64.0 & 1.1/0.4 \\
    TM-RCPN  & \textbf{82.3} & \textbf{79.1} & \textbf{64.5} & 1.6--6.1/0.9--5.9 \\
    \hline
  \end{tabular}
\end{table}

\subsection{SIFT Flow}
We report our results using multi-scale CNN features at three scales (1,1/2 and 1/4), as in \cite{rcpn}. Some of the classes in SIFT Flow dataset have a very small number of training instances, therefore, we also trained with balanced sampling to compensate for rare occurrence, referred to as \emph{bal.} prefix. We use 4 and 20 random trees for training and testing, respectively. The results for SIFT flow dataset are shown in Table~\ref{tab:SIFT-table}. PN-RCPN led to significant improvement in all three measures over RCPN and balanced training led to significant boost in MCA. The use of TM-RCPN does not affect the results much compared to PN-RCPN. We observe a strong trade-off between PPA and MCA on this dataset. Our overall best model in terms of both PPA and MCA (\emph{bal. TM-RCPN}) looks equivalent to the work in \cite{adobe}; PPA: 76.4 vs. 79.8, MCA: 52.6 vs. 48.8. 

\begin{table}[tb]
  \caption{SIFT Flow result.}
  \label{tab:SIFT-table}
  \centering
  \begin{tabular}{|c|c|c|c|c|}
    \hline
    \textbf{Method} & {\bf PPA} & {\bf MCA} & {\bf IoU} & \begin{tabular}{@{}c@{}}{\bf TPI (s)} \\ CPU/GPU\end{tabular}\\
    \hline\hline
    Tighe, \cite{lanaSIFT-IJCV}& 77.0 & 30.1 & NA & 8.4 / NA  \\
    Liu, \cite{liuSIFT}& 76.7 & NA & NA & 31 / NA  \\
    Singh, \cite{singhSIFT}& 79.2 & 33.8 & NA & 20 / NA \\
    Eigen, \cite{eigenSIFT} & 77.1 & 32.5 & NA & 16.6 / NA  \\
    Farabet, \cite{lecunSegmentation} & 78.5 & 29.6 & NA & NA / NA\\
    (Balanced), \cite{lecunSegmentation}& 72.3 & 50.8 & NA & NA / NA  \\
    Tighe, \cite{lanaSIFT2013}& 78.6 & 39.2 & NA & $\geq 8.4$ / NA  \\
    Pinheiro, \cite{JMLRseg} & 77.7 & 29.8 & NA & NA / NA \\
    Yang, \cite{adobe} & 79.8 & 48.7 & NA & $\leq 12$/NA \\
    \hline \hline
    Plain-NN &  76.3 & 32.1 & 24.7 & 1.1/0.36 \\
    \hline \hline
    RCPN, \cite{rcpn} & 79.6 & 33.6 & 26.9 & 1.1/0.4 \\
    bal. RCPN, \cite{rcpn}  & 75.5 & 48.0 & 28.6 & 1.1/0.4 \\
    \hline \hline
    PN-RCPN  & \textbf{80.9} & 39.1 & 30.8 & 1.1/0.4 \\
    bal. PN-RCPN  & 75.5 & \textbf{52.8} & 30.2 & 1.1/0.4 \\
    TM-RCPN & 80.8 & 38.4 &  30.7 & 1.6--6.1/0.9--5.4 \\
    bal. TM-RCPN   & 76.4 & 52.6 & \textbf{31.4} & 1.6--6.1/0.9--5.8 \\
    \hline
  \end{tabular}
\end{table}

\subsection{Daimler Urban}

\begin{figure}
\centering
      \includegraphics[keepaspectratio=false,width=8.5cm,height=5cm]{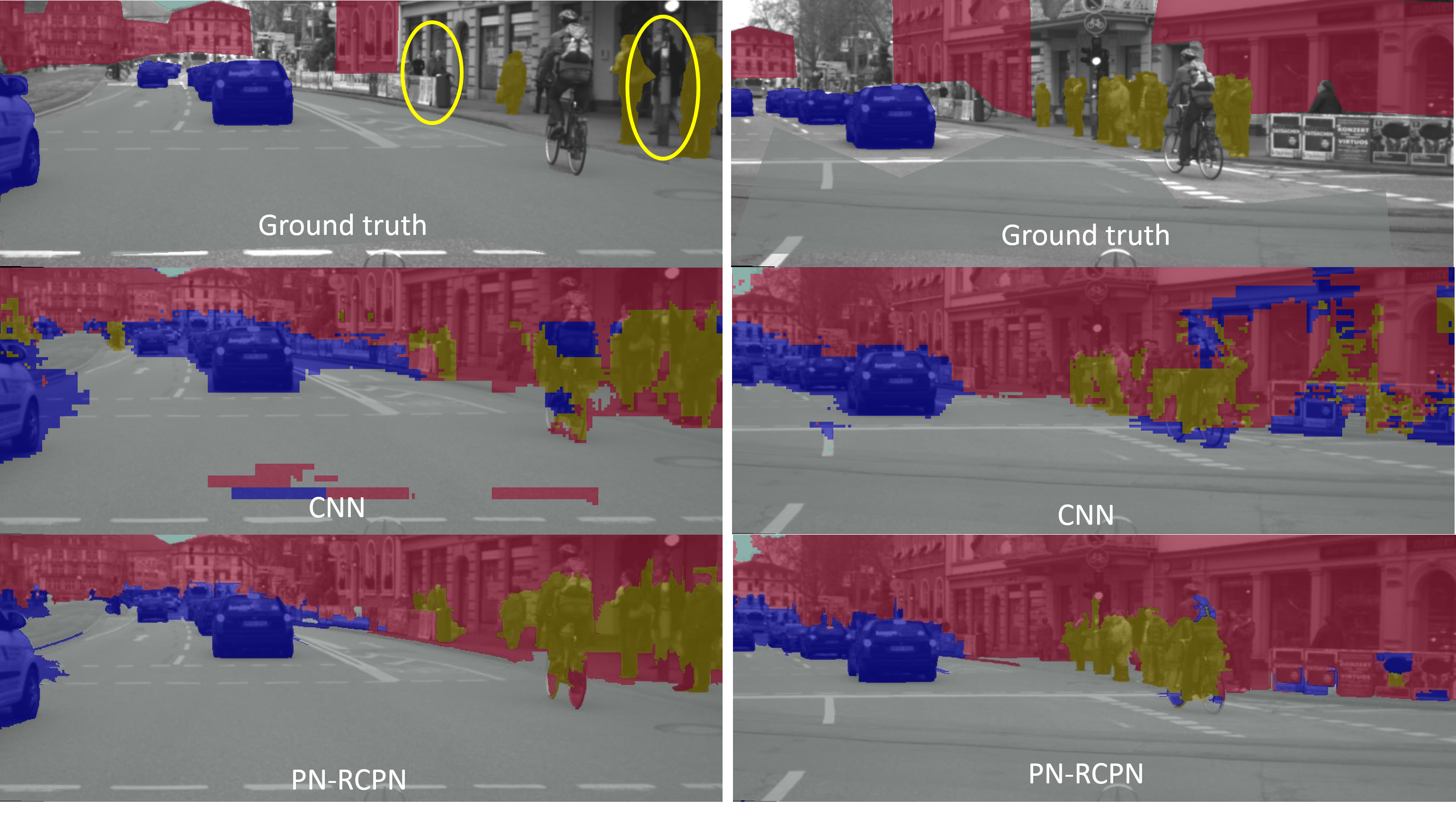}
      \caption{Some representative image segmentation results on Daimler Urban dataset. Here, CNN refers to direct per-pixel classification resulting from the multi-scale CNN. The ground-truth images are only partially labeled and we have shown the unlabeled pedestrians by yellow ellipses.}
      \label{fig:daimlerResult}
\end{figure}

We report our results using multi-scale CNN features with balanced training in Table~\ref{tab:daimler-table}. The previous results are based on the predicted labels provided by the authors of \cite{stixmantics}. The authors, in their paper \cite{stixmantics}, have reported the results with background as one of the classes, but the ground-truth labels for this dataset have portions of foreground classes labeled as the background. Therefore, even a correct segmentation is penalized. We ignore the background class while reporting the results for a fair evaluation. {\bf IoU Dyn} is the IoU for dynamic objects ie cars, pedestrians and bicyclists. We would like to underscore that the previous approaches (\cite{ale,stixmantics}) use stereo, depth, visual odometry and multi-frame temporal information that relies on the fact that the images are coming from a moving vehicle whereas, we only use an independent single visual image and still obtain similar or better performance. We observe significant improvements in terms of IoU with the use of PN-RCPN over RCPN and Plain-NN which could be due to the well structured image semantics of this dataset that allows it to learn the structure very effectively and utilize the context in a much better way than the other two datasets. Some of the representative segmentation results are shown in Fig.~\ref{fig:daimlerResult}. We have also submitted a complete video of semantic segmentation for all the test images for Daimler urban in the supplementary material.

\subsection{Segmentation Time}
In this section we provide the timing details for the experiments. Only the Multi-CNN feature extraction is executed on a GPU for our Plain-NN and RCPN variants. Due to similar image sizes, SIFT flow and Stanford Background took almost the same computation per image except while using TM-RCPN, because of the difference in label state-space size. The time break-up for SIFT flow (same for Stanford) in seconds is 0.3 (super-pixellation) + 0.08/0.8 (GPU/CPU visual feature) + 0.01 (PN-RCPN) + 0.5--5 (TM-MRF). For Daimler, the corresponding timings are 2.4 + 0.4/3.5 + 0.09 + 6 seconds. Therefore, the bottleneck for our system is the super-pixellation time for PN-RCPN and MRF inference for TM-RCPN. Fortunately, there are real-time super-pixellation algorithms, such as \cite{pedro}, that can help us achieve state-of-the-art semantic segmentation within 100 milliseconds on an NVIDIA Titan Black GPU.

\begin{table}[tb]
  \caption{Daimler result. Numbers in {\it italics} indicate the use of stereo, depth and multi-frame temporal information.}
  \label{tab:daimler-table}
  \centering
  \begin{tabular}{|c|c|c|c|c|c|}
    \hline
    \textbf{Method} & {\bf PPA} & {\bf MCA} & {\bf IoU} & {\bf IoU Dyn} & \begin{tabular}{@{}c@{}}{\bf TPI (s)} \\ CPU/GPU\end{tabular}\\
    \hline\hline
    Joint, \cite{ale,stixmantics} & \textit{\textbf{ 94.5}} & \textbf{\textit{91.0}} & \textbf{\textit{86.0}} & \textbf{\textit{74.5}}  & \textit{111 / NA}  \\
    Stix., \cite{stixmantics} & \textit{92.8} & \textit{87.5} & \textit{80.6}  & \textit{72.3} & \textit{\textbf{0.05} / NA}  \\
    \hline \hline
    bal. Plain-NN & 91.4 &  83.2 & 75.8 &  56.2 & 5.9 / 2.8  \\
    \hline \hline
    bal. RCPN & 93.3     &   87.6    &    80.9    &    66.0 & 6.0 / 2.8  \\
    \hline \hline
    bal. PN-RCPN& \textbf{94.5}   &     90.2  &      84.5  &      73.8  & 6.0 / 2.8  \\
    bal. TM-RCPN& \textbf{94.5}   &     90.1  &      84.5  &      73.8  & 12 / 8.8  \\
    \hline
  \end{tabular}
\end{table}

\section{Conclusion}\label{sec:conclusion}
We analyzed the recursive contextual propagation network, referred to as RCPN \cite{rcpn} and discovered potential problems with the learning of it's parameters. Specifically, we showed the existence of bypass errors and explained how it can reduce the RCPN model to an effective multi-layer neural network for each super-pixel. Based on our findings, we proposed to include the classification loss of pure-nodes to the original RCPN formulation and demonstrated it's benefits in terms of avoiding the bypass errors. We also proposed a tree MRF on the parse tree nodes to utilize the pure-node's label estimation for inferring the super-pixel labels. The proposed approaches lead to state-of-the-art performance on three segmentation datasets: Stanford background, SIFT flow and Daimler urban.

{\small
\bibliographystyle{ieee}
\bibliography{nips2014}
}

\end{document}